%% file: lim2023ur.tex
\documentclass[a4paper, 10 pt, conference]{URL-ieeeconf}
\IEEEoverridecommandlockouts    
\input{URL-latex}

\input{URL-math}
\title{\LARGE \bf AdaLIO: Robust Adaptive LiDAR-Inertial Odometry in \\ Degenerate Indoor Environments}

\author{Hyungtae Lim$^{1\dagger}$, Daebeom Kim$^{1\dagger}$, Beomsoo Kim$^{2}$, and Hyun Myung$^{1*}$, \textit{Senior Member, IEEE}
  \thanks{$^*$Corresponding author: Hyun Myung}
  \thanks{$^\dagger$These authors contributed equally to this work}
  \thanks{$^{1}$Hyungtae Lim, Daebeom Kim, and Hyun Myung are with the School of Electrical Engineering, KAIST (Korea Advanced Institute of Science and Technology), Daejeon, 34141, Republic of Korea. {\tt\scriptsize \{shapelim, ted97k, hmyung\}@kaist.ac.kr} \hfill \break   	
  \indent $^{2}$Beomsoo Kim was with the School of Electrical Engineering, KAIST (Korea Advanced Institute of Science and Technology), Daejeon, 34141, Republic of Korea as a research intern. {\tt\scriptsize qjatn68@gmail.com} \hfill \break   	
  \indent This work was supported in part by Korea Evaluation Institute of Industrial Technology~(KEIT) funded by the Korea Government (MOTIE) under Grant No.20018216, Development of mobile intelligence SW for autonomous navigation of legged robots in dynamic and atypical environments for real application. The students are supported by Korea Ministry of Land, Infrastructure and Transport (MOLIT) as ``Innovative Talent Education Program for Smart City'' and BK21 FOUR~(Republic of Korea).}
}

\begin{document}
\maketitle
\thispagestyle{empty}
\pagestyle{empty}

\begin{abstract}
  %
 In recent years, the demand for mapping construction sites or buildings using light detection and ranging~(LiDAR) sensors has been increased to model environments for efficient site management.  
 However, it is observed that sometimes LiDAR-based approaches diverge in narrow and confined environments, such as spiral stairs and corridors, caused by fixed parameters regardless of the changes in the environments. 
 That is, the parameters of LiDAR (-inertial) odometry are mostly set for open space; thus, if the same parameters suitable for the open space are applied in a corridor-like scene, it results in divergence of odometry methods, which is referred to as \textit{degeneracy}.
 To tackle this degeneracy problem, we propose a robust LiDAR inertial odometry called \textit{AdaLIO}, which employs an adaptive parameter setting strategy. To this end, we first check the degeneracy by checking whether the surroundings are corridor-like environments.
 If so, the parameters relevant to voxelization and normal vector estimation are adaptively changed to increase the number of correspondences. 
 As verified in a public dataset, our proposed method showed promising performance in narrow and cramped environments, avoiding the degeneracy problem.

\end{abstract}

\section{Introduction}
\label{sec:intro}

In recent years, the demand for mapping construction sites or buildings has been increased to model environments~\cite{zhang2022ral,zhang2014rss, lim2020normal, lim2021ral,ren2019sensors}. 
This mapping presents the geometrical information for inspection robots~\cite{jung2020rs} or surveyors~\cite{sung2021isr}, which enables us to achieve robotic navigation or survey, respectively.
Thus, mapping the construction sites or buildings allows efficient site management.

Various sensors are used to model the surroundings~\cite{song2022ral,kim2022ral-step,noh2023ral,hu2022iros,lee2022ral-vivid}. Among them, light detection and ranging~(LiDAR) sensors are widely utilized~\cite{zhang2022ral} because LiDAR sensors can acquire centimeter-wise accuracy measurements based on laser scans~\cite{lim2021ral-patch, seo2022ur}. Even though some sensors show better performance than LiDAR sensors in bad weather condition~\cite{burnett2021ral}, mapping of construction sites or buildings is usually conducted in indoor environments or when it does not snow or rain. Therefore, the accurate mapping characteristics of the LiDAR sensors are still valid in construction sites. 

To achieve precise mapping of surroundings, pose estimation is necessary. In general, this pose estimation can be classified into two sub-groups: one is LiDAR odometry~\cite{seo2022ur}, which estimates the pose of consecutive frames, and the other is registration, which estimates the relative pose between two point clouds. 
In LiDAR odometry, correspondence estimation between previous and current scan data is usually performed by nearest neighbor search~\cite{vizzo2023ral}.
These correspondences describe the inter-spatial geometrical relationship between the previously observed measurements and current measurement. Therefore, the quality of data association between the consecutive frames directly affects mapping precision~\cite{zhang2014rss, vizzo2023ral}.

\begin{figure}[t!]
	\centering
	\begin{subfigure}[b]{0.23\textwidth}
		\includegraphics[width=1.0\textwidth]{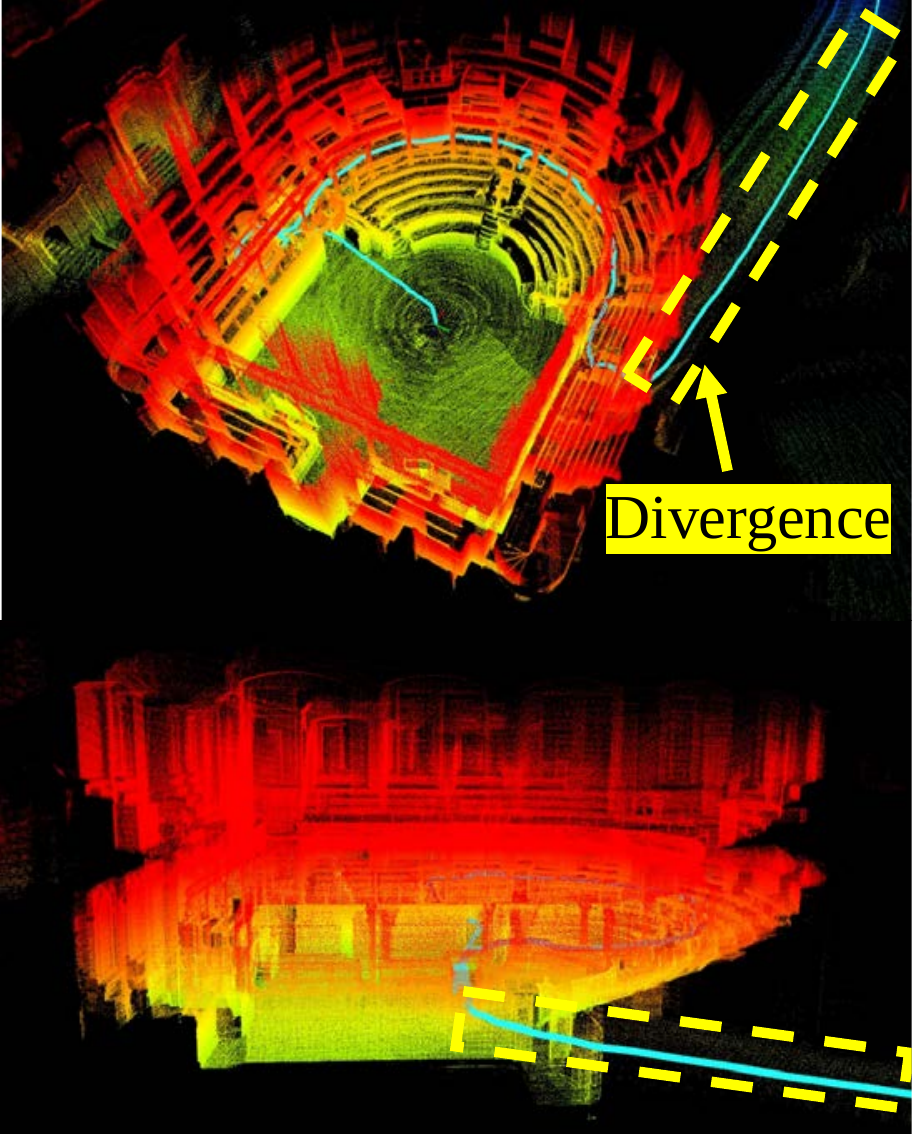}
		\caption{Faster-LIO~\cite{bai2022ral}}
	\end{subfigure}
	\begin{subfigure}[b]{0.23\textwidth}
		\includegraphics[width=1.0\textwidth]{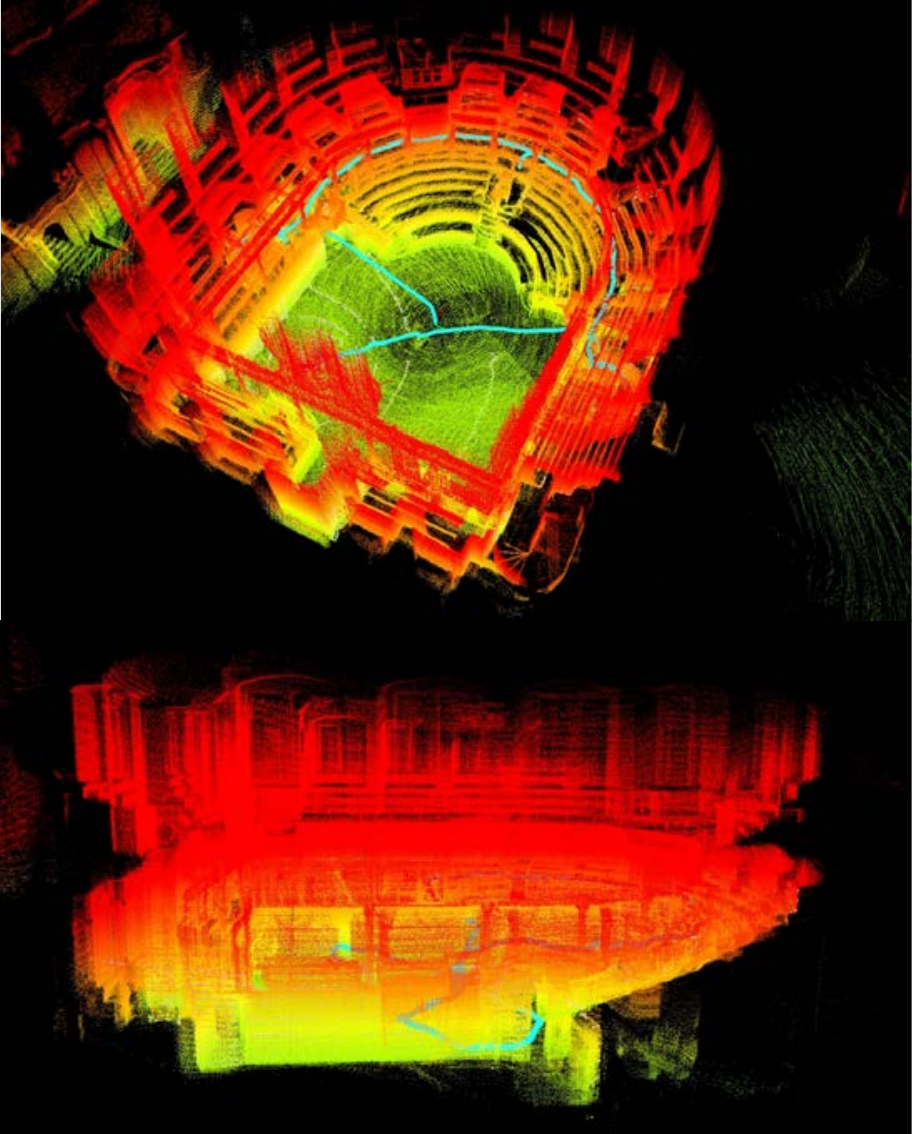}
		\caption{AdaLIO~(Ours)}
	\end{subfigure}
	\caption{Performance comparison of (a)~Faster-LIO~\cite{bai2022ral} and (b)~our proposed LiDAR-inertial odometry, called \textit{AdaLIO}, on \texttt{Exp11} of the HILTI-Oxford dataset~\cite{zhang2022ral}. By exploiting an adaptive parameter setting strategy depending on whether the surroundings are corridor-like environments or not, our AdaLIO can robustly estimate ego-motion in a narrow environment~(best viewed in color).}
	\label{fig:fig1}
\end{figure}

However, it is observed that sometimes LiDAR-based approaches diverge in narrow and cramped environments, such as spiral stairs and corridors, caused by fixed parameters regardless of the changes in the environments, as shown in~\figref{fig:fig1}(a). 
That is, the parameters of LiDAR (-inertial) odometry are mostly set for open space; thus, if the same parameters that are suitable for the open space are applied in a corridor-like scene, it results in divergence of odometry methods.
For instance, the voxel size of voxelization is empirically set to range between 0.2\;m and 0.5\;m to for efficient representation of scan data by abstracting some points in the same voxel into a single point, which makes the number of points become too few in these narrow scenes. 
Consequently, the number and quality of the correspondence become degraded. 
In this paper, we call this phenomenon \textit{degeneracy}.
Therefore, a robust LiDAR~(-inertial) odometry method is required to achieve precise indoor mapping.

To tackle this degeneracy problem, we propose a robust LiDAR inertial odometry called \textit{AdaLIO}, which employs an adaptive parameter setting strategy. To this end, we first check the degeneracy by checking whether the surroundings are corridor-like environments. 
Then, if the surroundings are highly likely to be narrow and cramped scenes, the adaptive parameter setting strategy is applied to let the number of correspondences increase. By doing so, our proposed method can avoid divergence in the narrow and closed scenes~(\figref{fig:fig1}(b)).

In summary, the contribution of this paper is threefold:

\begin{itemize}

\item We propose a degenerate-environment-robust LIO to prevent divergence in cramped environments by avoiding degeneracy.

\item To this end, the adaptive parameter setting strategy, which adaptively adjusts some parameters regarding voxelization and correspondence estimation, is proposed to preserve the number of correspondences, making our approach robust against these narrow scenes. 

\item In qualitative evaluation, our AdaLIO showed promising performance compared with a state-of-the-art method~\cite{bai2022ral} in the narrow and cramped environments.
\end{itemize}


\section{Related Work}
\label{sec:related}


\subsection{LiDAR (-Inertial) Odometry}

There are two main categories in LiDAR~(-inertial) odometry fields: one is indirect methods, which extract features from the point cloud and the other is direct methods, which use the entire point cloud for estimating poses after voxelization.
Zhang~\etalcite{zhang2014rss} proposed LOAM, a real-time method for odometry and mapping, which extracts edge and planar features. 
Shan~\etalcite{shan2018iros} proposed a lightweight and ground-optimized lidar odometry and mapping method, LeGO-LOAM, which exploits ground information to make the pose estimation more robust and efficient.

In cases of the direct methods, Xu~\etalcite{xu2021ral} proposed a fast and robust LiDAR-inertial odometry, Fast-LIO, which is based on a tightly-coupled iterated extended Kalman filter. Vizzo~\etalcite{vizzo2023ral} proposed an accurate and robust odometry estimation based on point-to-point ICP with a robust kernel.
Hinduja~\etalcite{hinduja2019iros} proposed the degeneracy-aware point-to-plane ICP algorithm by optimizing the parameters through well-constrained directional space when the degeneracy is detected.

\subsection{LiDAR Feature Extraction}

There are some efforts to solve the geometric degeneracy problem only with the information obtained by a LiDAR sensor. Li~\etalcite{li2022tim} proposed an intensity-augmented LiDAR-inertial SLAM in degenerated environments by additionally extracting planar points and intensity edge points. 
Du~\etalcite{du2023arxiv} proposed a novel LiDAR intensity image-based SLAM method for robustly extracting features from intensity images and prune unnecessary features. 
Jiao~\etalcite{jiao2021icra} proposed a novel adaptive feature selection method to be robust against degenerate environments.

\subsection{Sensor Fusion}

Recently, LiDAR sensors have been commonly fused with additional sensors to solve the degeneracy problem. Shan~\etalcite{shan2021icra} proposed tightly-coupled LiDAR-visual-inertial odometry via smoothing and mapping, called LVI-SAM. LVI-SAM utilizes visual features and initial guesses obtained using cameras and IMU to make the ego-motion estimation robust in feature-less and texture-less environments. 
Some researchers have proposed fusion methods leveraging an ultra-wideband (UWB) sensor because the range measurements from the UWB sensors are less affected by influences of the degeneracy. 
For this reason, Zhen~\etalcite{zhen2019icra} proposed a novel localizability estimation method and Zhou~\etalcite{zhou2021tvt} showed the feasibility of a LiDAR sensor and UWB fusion algorithm, aiming for SLAM with anti-degeneration capability. 

\begin{figure}[t!]
	\centering
	\begin{subfigure}[b]{0.45\textwidth}
		\includegraphics[width=1\textwidth]{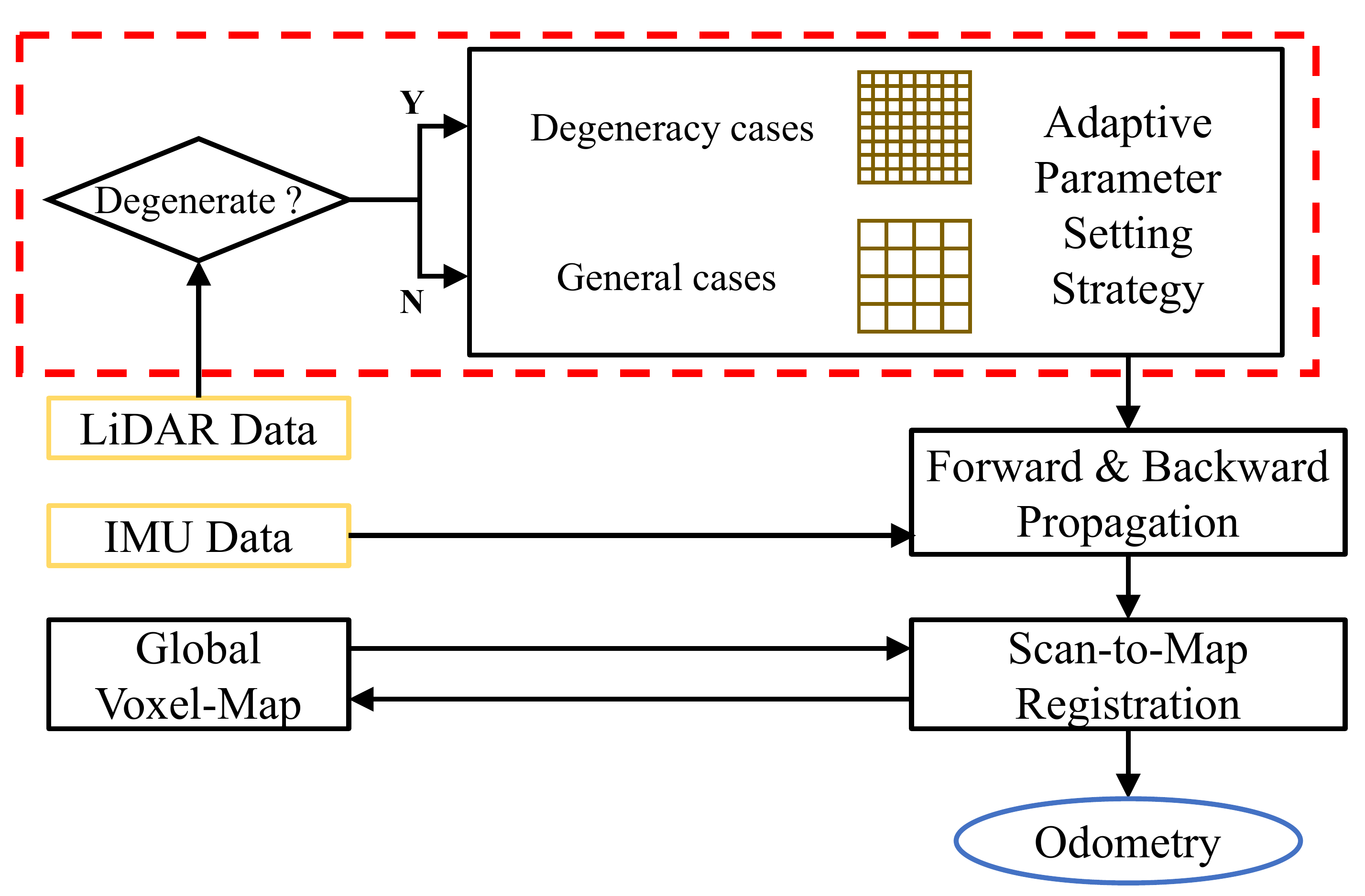}
	\end{subfigure}
	\caption{System overview of our robust adaptive LiDAR inertial odometry, called ~\textit{AdaLIO}.}
	\label{fig:AdaLIO_pipeline}
\end{figure}

\section{AdaLIO: Robust LiDAR Inertial Odometry Leveraging Adaptive Parameter Setting}

In this paper, we propose a novel adaptive LiDAR-Inertial odometry method that adaptively adjusts internal parameters by checking degeneracy. That is, if the degeneracy is detected, our proposed method increases the number of correspondences by changing parameters relevant to the data association.

\subsection{Problem Definition}

We present a problem definition to tackle the divergence of odometry in a degenerate environment, which is shown in~\figref{fig:fig1}(a).
We observe that the divergence occurs once the number of correspondences of Faster-LIO is drastically decreased due to the parameters that are mostly suitable for open space.
Furthermore, the voxelization in the narrow scenes also degrades an imbalance of the distribution of normal vectors. 
That is, normal vector estimation in a submap is followed by this voxelization for calculating point-to-plane residuals between the submap and current scan at time step $t$. 
However, in degenerate environments, normal vectors are extracted only from the partial walls and located close together.

Therefore, we tackle this problem in the following two steps. First, we check whether the surrounding is likely to be degenerate scenes. If so, the parameters relevant to the voxelization and normal vector estimation are set to increase the number of correspondences. More details are explained in the following paragraphs.
    
\subsection{System Overview}

System overview is shown in~\figref{fig:AdaLIO_pipeline}.
Our method is based on Faster-LIO~\cite{bai2022ral}, which uses an iterated error state Kalman filter on the 3D manifold to estimate the relative pose based on forward and backward propagation. To estimate correspondences and set residuals between the submap and current scan, Faster-LIO utilizes the nearest neighbor search and point-to-plane distance, respectively. 
Our approach contains motion distortion compensation through the forward and backward propagation steps by using IMU data. Then, the relative pose is estimated through scan-to-map registration and then a global voxel map is updated.

The red dashed block in~\figref{fig:AdaLIO_pipeline} indicates our proposed approach. By checking whether the observed surroundings are narrow scenes, the adaptive parameter strategy is applied to adaptively set the parameters. 

\subsection{Adaptive Parameter Setting Strategy}

To be more concrete, we change parameters relevant to normal vector estimation, which are used for calculating point-to-plane distances, and voxelization. We describe how adjusting those parameters can solve the degeneracy.


First, fixed voxel size for voxelization regardless of the changes in the environments is likely to abstract the current scan too much. Voxelization represents some points in the same voxel into a single point, so the number of points becomes fewer in these narrow scenes than in open space. This is because the difference in the measured volume is quite large. To tackle this problem, if the number of voxel-sampled cloud points is smaller than in the general case and most occupied voxels are close to the origin of the sensor frame, adaptive voxelization is applied. Through this approach, voxelization is conducted with a smaller voxel size, increasing the number of voxelized cloud points.

Second, to forcibly preserve the number of measurements, the parameters for estimating normal vector and setting the point-to-plane distance are also adaptively adjusted. In a degenerate environment, the distribution of normal vectors is also imbalanced~\cite{park2017iccv, kang2016tro}. To resolve that issue, the parameters relevant to this procedure are changed depending on the surroundings. Thus, the range of the search radius and the absolute residual margin of the estimated plane are set as smaller values. 
The search radius denotes the distance to find the nearest points of a given query point from the current scan. The residual margin denotes the margin for checking whether the estimated normal vector is reliable. These parameters are interrelated, so if the search radius is set to be a smaller value, then the residual margin should also be set as a smaller value to get the valid and reliable normal vector and point-to-plane distance~\cite{wang2017tmi},~\cite{schaffert2019tmi}. 
Note that a smaller residual margin means that the normal vector is checked more conservatively to reject potential outlier correspondences caused by the smaller search radius.

This adaptive setting is simple, yet allows better matching of correspondences and finally prevents divergence of odometry~(see \secref{subsec:qualitative} and \secref{subsec:quantitative}).

\section{Experimental Evaluation}
\label{sec:exp}

The main focus of this work is a degenerate-environment-robust LIO approach against narrow and cramped environments by leveraging the adaptive parameter setting strategy. We present our experiments to show the robustness of our method. 

\subsection{Experimental Setup}

To perform our analysis, we exploit the HILTI-Oxford dataset~\cite{zhang2022ral}, which is also utilized in the HILTI-Oxford SLAM Challenge 2022\footnote{https://hilti-challenge.com}. Originally, the dataset consists of more than ten sequences, but most of their ground truth poses are not provided because these sequences are originally for the challenge. For this reason, we only use \texttt{Exp01}, \texttt{Exp02}, \texttt{Exp03}, \texttt{Exp04}, \texttt{Exp05}, and \texttt{Exp06}, whose ground poses are available as a validation dataset. In addition, we employ \texttt{Exp11} for qualitative comparison because \texttt{Exp11} includes very narrow spiral stairs.

For quantitative evaluation, the HILTI-Oxford dataset provides millimeter-level marker poses for each sequence, but these marker poses are not full trajectories. Thus, if the distance between a marker pose and the closest pose from the estimated trajectory satisfies $\leq$ 1\;cm, $\leq$ 10\;cm, or $\leq$ 100\;cm, we score that case as 10, 6, or 3, respectively. See Zhang~\etalcite{zhang2022ral} for more details on these metrics.

Our parameters for adaptive parameter setting strategy are summarized in Table~\ref{table:param}.

\begin{table}[t!]
	\centering
	\caption{Parameter settings of our proposed adaptive parameter setting strategy~(unit: m).}
	\setlength{\tabcolsep}{4pt}
	\begin{tabular}{l|cccccc}
		\toprule \midrule
		Parameters &  In general cases & In degeneracy cases  \\ \midrule
		Voxelization size & 0.2 & 0.1  \\
		Search radius & 3.0 & 2.0  \\
		Residual margin of the plane & 0.05 & 0.025  \\
		\midrule\bottomrule
	\end{tabular}
	\label{table:param}
\end{table}

\begin{figure*}[t!]
	\centering
	\begin{subfigure}[b]{1.0\textwidth}
		\includegraphics[width=1.0\textwidth]{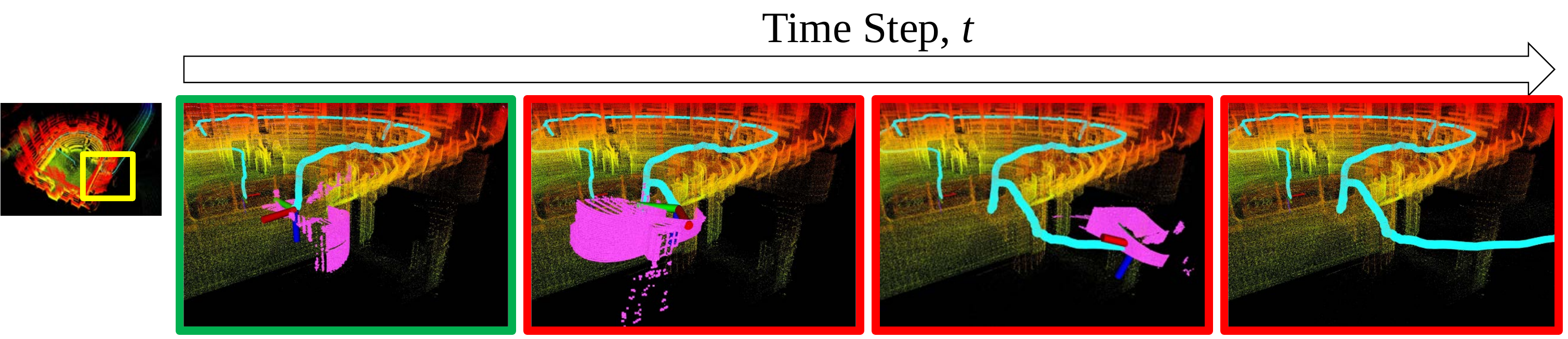}
	\end{subfigure}
	\begin{subfigure}[b]{1.0\textwidth}
		\includegraphics[width=1.0\textwidth]{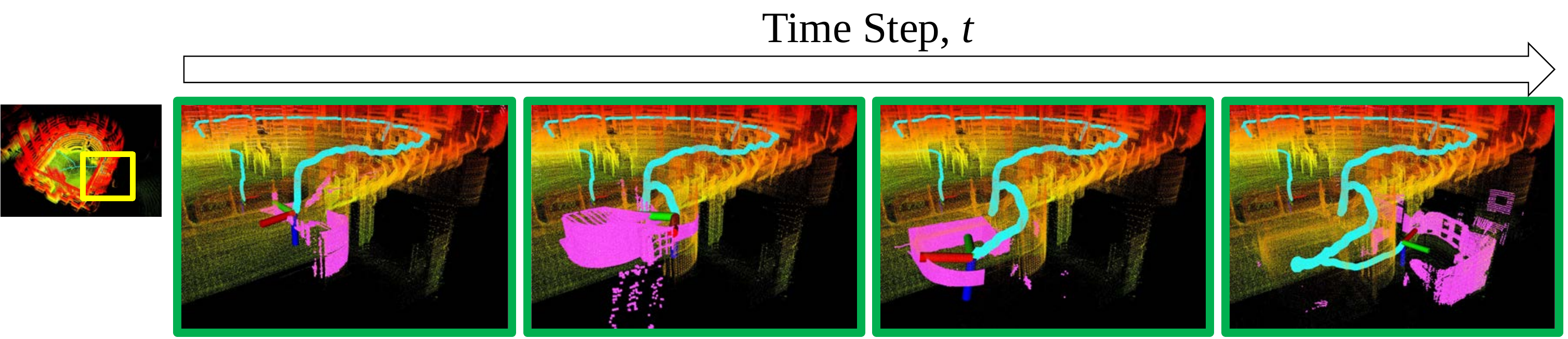}
	\end{subfigure}
	\caption{(T-B): Qualitative comparison of Faster-LIO~\cite{bai2022ral} and our proposed method. The red and green frame indicate that an algorithm diverges and successfully estimate the relative pose, respectively}
	\label{fig:qualitative}
\end{figure*}

\begin{figure*}[t!]
	\centering
	\begin{subfigure}[b]{0.40\textwidth}
		\includegraphics[width=1\textwidth]{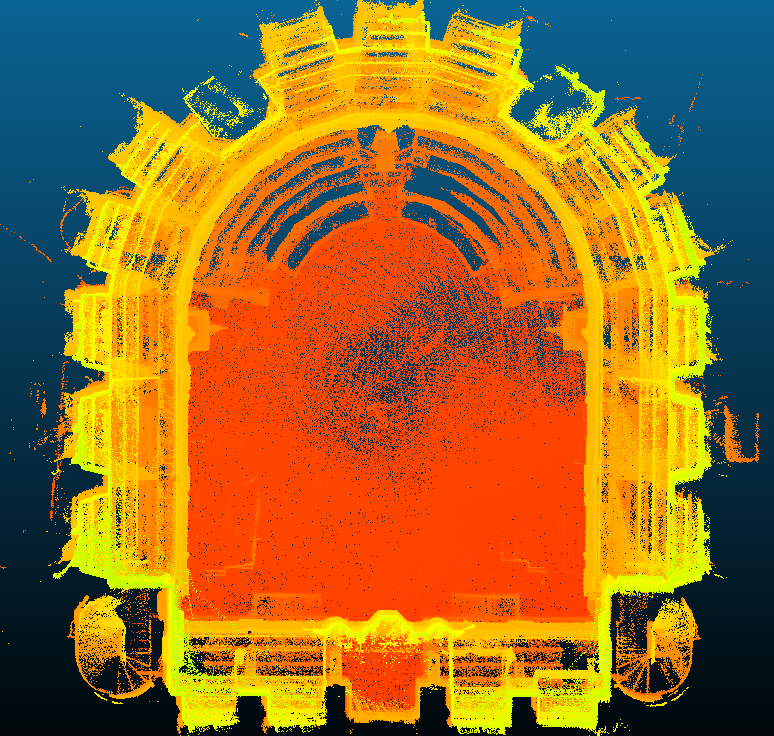}
		\caption{}
	\end{subfigure}
	\begin{subfigure}[b]{0.538\textwidth}
		\includegraphics[width=1\textwidth]{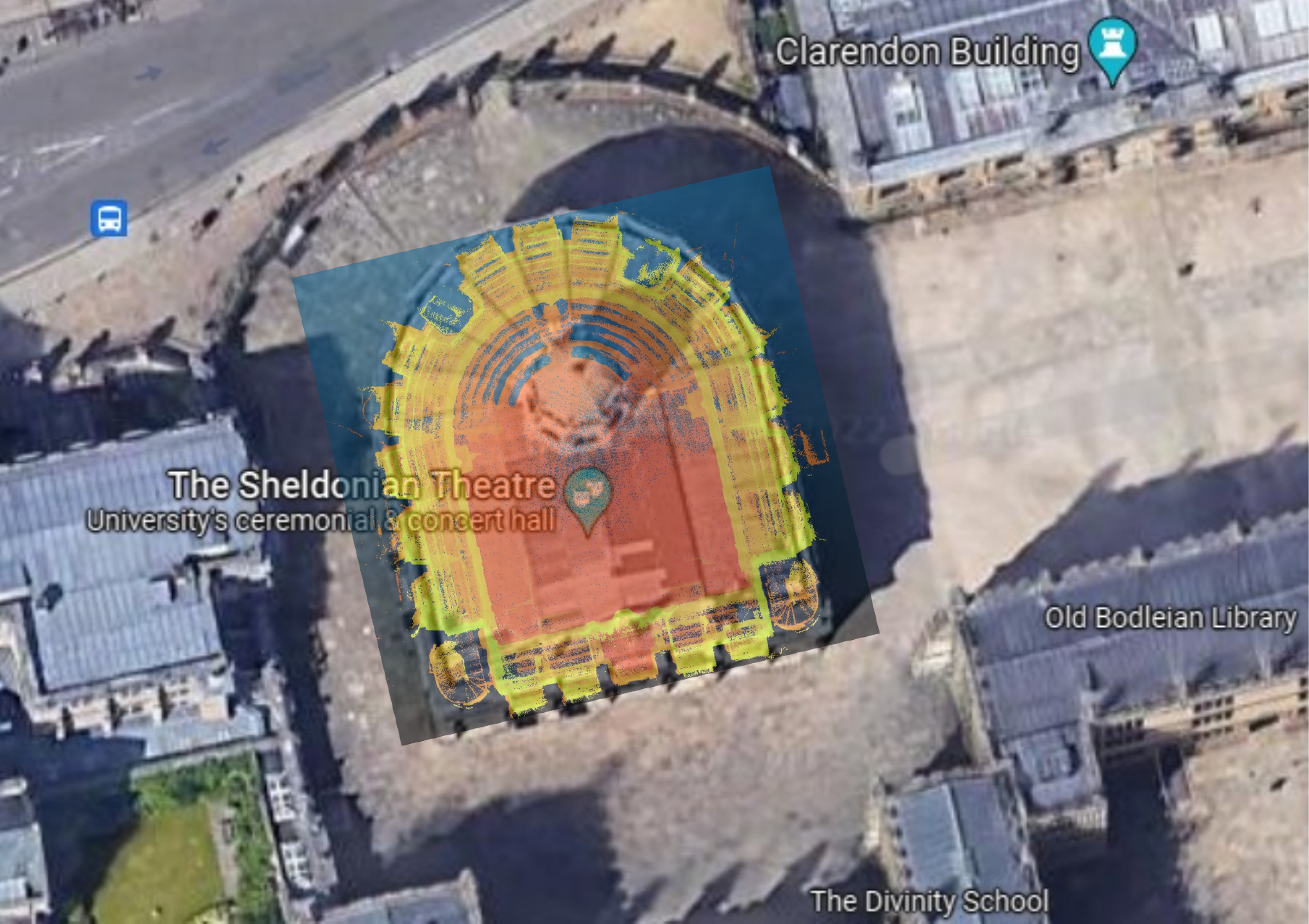}
		\caption{}
	\end{subfigure}
	\caption{(a) Visualized result of mapping in \texttt{Exp11} in HILTI-Oxford dataset and (b)~visualized map overlay on Google Maps~(best viewed in color).}
	\label{fig:sheldonian}
\end{figure*}

\subsection{Qualitative Performance Comparison In Degenerate Environments}\label{subsec:qualitative}

First, we qualitatively compared our proposed algorithm with Faster-LIO in terms of robustness in \texttt{Exp11}. As shown in~\figref{fig:qualitative}, when descending a narrow and tight spiral staircase, Faster-LIO failed to estimate pose. This is because Faster-LIO uses a fixed voxel size, resulting in a drastic decrease in the number of correspondences between the previously accumulated submap and the current point cloud. For this reason, Faster-LIO failed to pass through the narrow spiral staircase and thus diverged.

In contrast, our proposed algorithm successfully estimated poses even though a surveyor descends the spiral staircase. By adaptively adjusting the parameters, our AdaLIO detected the narrow staircase and forcibly increased the number of correspondences. Therefore, AdaLIO could prevent divergence.

\subsection{Quantitative Performance Comparison}\label{subsec:quantitative}

Additionally, we conduct a quantitative evaluation of two algorithms. As shown in \tabref{table:score}, our approach achieved a higher total score. This is because the adaptive parameter setting strategy in our AdaLIO enables to estimate the pose more accurately, preventing divergence even in cases where Faster-LIO fails to estimate the pose due to the divergence. 

\begin{figure}[t!]
	\centering
	\begin{subfigure}[b]{0.23\textwidth}
		\includegraphics[width=1.0\textwidth]{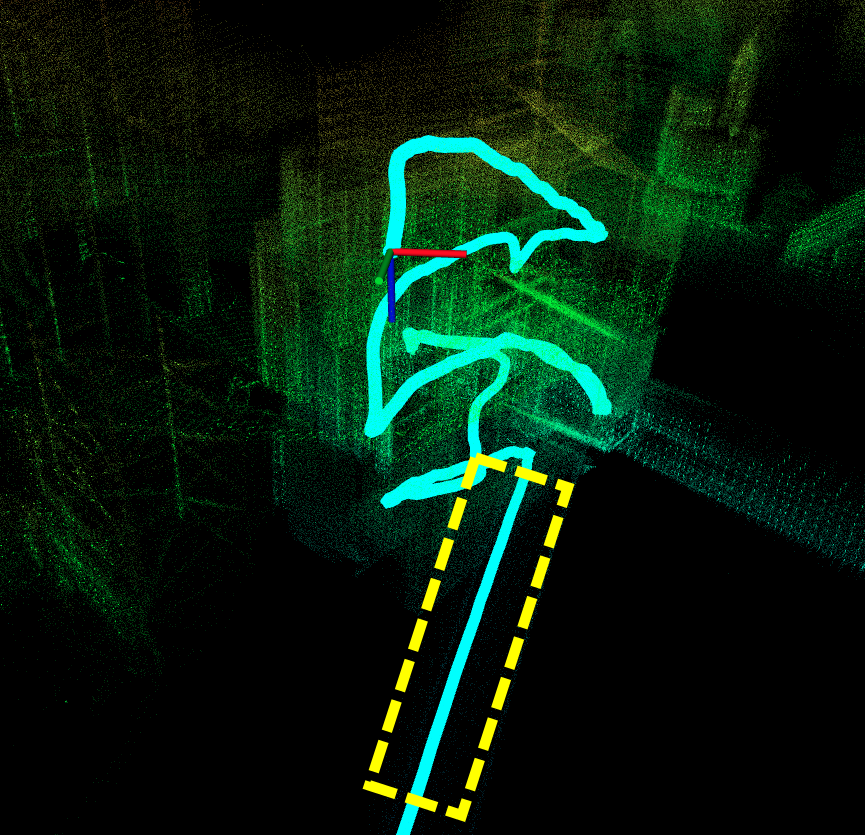}
		\caption{Faster-LIO~\cite{bai2022ral}}
	\end{subfigure}
	\begin{subfigure}[b]{0.23\textwidth}
		\includegraphics[width=1.0\textwidth]{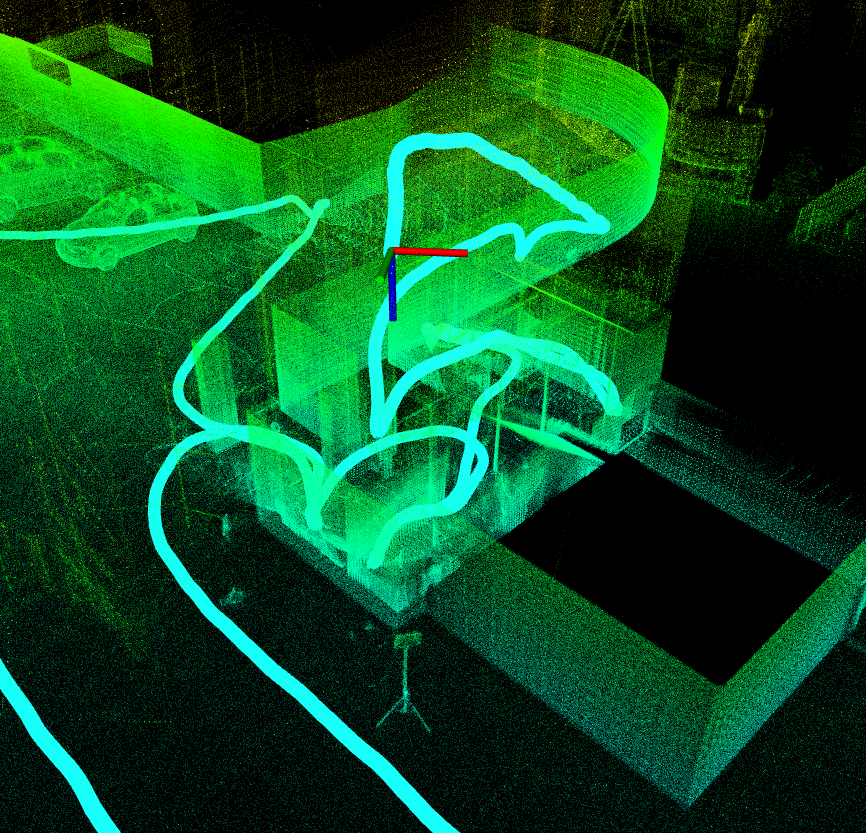}
		\caption{AdaLIO~(Ours)}
	\end{subfigure}
	\caption{Qualitative comparison of Faster-LIO~\cite{bai2022ral} and our proposed method in \texttt{Exp03}.}
	\label{fig:exp03}
\end{figure}

Because our proposed method is also based on Faster-LIO, there was no significant performance difference in \texttt{Exp02} and \texttt{Exp04}, which do not contain narrow space.
However, a significant difference was observed in \texttt{Exp3}, as shown in~\figref{fig:exp03}. In \texttt{Exp3}, there was a section where a surveyor descended a narrow staircase and moved towards an underground parking lot. Under that circumstance, while Faster-LIO diverged when descending the narrow staircase~(\figref{fig:exp03}(a)), our proposed algorithm robustly estimated the pose without divergence~(\figref{fig:exp03}(b)). Therefore, our proposed algorithm obtained a significantly higher score compared to Faster-LIO.

In conclusion, all the experimental results support our claims that our proposed algorithm successfully prevents divergence by exploiting the adaptive parameter setting strategy.

\begin{table}[t!]
	\centering
	\caption{Reported scores of Faster-LIO~\cite{bai2022ral} and our AdaLIO in validation set of the HILTI SLAM Challenge 2022. $\times$ indicates the trajectory of the algorithm diverges.}
	\setlength{\tabcolsep}{4pt}
	\begin{tabular}{ll|ccc|c}
		\toprule \midrule
		Seq. & Method & $\leq$ 1cm & $\leq$ 10 cm & $\leq$ 100 cm &  Score  \\ \midrule
		\multirow{2}{*}{\texttt{Exp01}} & Faster-LIO~\cite{bai2022ral} &  0 & 8 & 5 & 63 \\
		& AdaLIO~(Ours) &  0 & 9 & 4 & \textbf{66} \\ \midrule
		\multirow{2}{*}{\texttt{Exp02}} & Faster-LIO~\cite{bai2022ral} & 0 & 12 & 10 & \textbf{102} \\
		& AdaLIO~(Ours) & 0 & 12 & 10 & \textbf{102} \\ \midrule
		\multirow{2}{*}{\texttt{Exp03}} & Faster-LIO~\cite{bai2022ral} & $\times$ & $\times$  & $\times$ & $\times$ \\ 
		& AdaLIO~(Ours) & 0 & 10 & 7 & \textbf{81} \\ \midrule
		\multirow{2}{*}{\texttt{Exp04}} & Faster-LIO~\cite{bai2022ral} & 0 & 6 & 1 & \textbf{39} \\
		& AdaLIO~(Ours) & 0 & 6 & 1 & \textbf{39} \\ \midrule
		\multirow{2}{*}{\texttt{Exp05}} & Faster-LIO~\cite{bai2022ral} & 0 & 4 & 2 & 30 \\
		& AdaLIO~(Ours) & 0 & 5 & 1 & \textbf{33} \\ \midrule
		\multirow{2}{*}{\texttt{Exp06}} & Faster-LIO~\cite{bai2022ral} & 0 & 3 & 4 & 30 \\
		& AdaLIO~(Ours) & 0 & 4 & 3 & \textbf{33} \\ \midrule \midrule  
		\multirow{2}{*}{Total} & Faster-LIO~\cite{bai2022ral} & 0 & 33 & 22  & 264 \\
					& AdaLIO~(Ours)  & 0 & 46 & 26 & \textbf{354} \\
		\midrule \bottomrule
	\end{tabular}
	\label{table:score}
\end{table}

\section{Conclusion}
\label{sec:conclusion}

In this study, we presented a novel approach to LiDAR inertial odometry called \textit{AdaLIO}.
Our proposed adaptive parameter setting strategy in AdaLIO prevents degeneracy in a narrow and confined environment by changing parameters relevant to the data association.
We implemented and evaluated our approach on the HILTI-Oxford dataset. And we provided comparisons to other existing techniques, supporting all claims made in this paper.
In future works, we plan to propose a degeneracy-robust SLAM framework and test our proposed method in various robot platforms.



\bibliographystyle{URL-IEEEtrans}

\bibliography{URL-bib}

\end{document}

%% file: URL-latex.tex
\usepackage{graphics}           
\usepackage{times}              
\usepackage{amsmath}            
\usepackage{amssymb}            
\usepackage{graphicx}
\usepackage{algorithm}
\usepackage[noend]{algpseudocode}
\usepackage{booktabs}
\usepackage{color}
\usepackage{multirow}
\usepackage{subcaption}
\usepackage{rotating}
\usepackage{cite}
\definecolor{instructioncolor}{rgb}{.0,.2,.9}

\def\secref#1{Section~\ref{#1}}
\def\figref#1{Fig.~\ref{#1}}
\def\tabref#1{Table~\ref{#1}}
\def\eqref#1{(\ref{#1})}

\newcommand{\rom}[1]{\uppercase\expandafter{\romannumeral #1\relax}}

\makeatletter
\usepackage{xspace}
\DeclareRobustCommand\onedot{\futurelet\@let@token\@onedot}
\def\@onedot{\ifx\@let@token.\else.\null\fi\xspace}

\def\etal{{\textit{et~al}}\onedot}
\makeatother

\def\etalcite#1{\etal~\cite{#1}}

\usepackage{array}
\newcolumntype{L}[1]{>{\raggedright\let\newline\\\arraybackslash\hspace{0pt}}m{#1}}
\newcolumntype{C}[1]{>{\centering\let\newline\\\arraybackslash\hspace{0pt}}m{#1}}
\newcolumntype{R}[1]{>{\raggedleft\let\newline\\\arraybackslash\hspace{0pt}}m{#1}}

%% file: lim2023ur.bbl
\begin{thebibliography}{10}
\providecommand{\url}[1]{#1}
\csname url@rmstyle\endcsname
\providecommand{\newblock}{\relax}
\providecommand{\bibinfo}[2]{#2}
\providecommand\BIBentrySTDinterwordspacing{\spaceskip=0pt\relax}
\providecommand\BIBentryALTinterwordstretchfactor{4}
\providecommand\BIBentryALTinterwordspacing{\spaceskip=\fontdimen2\font plus
\BIBentryALTinterwordstretchfactor\fontdimen3\font minus
  \fontdimen4\font\relax}
\providecommand\BIBforeignlanguage[2]{{%
\expandafter\ifx\csname l@#1\endcsname\relax
\typeout{** WARNING: IEEEtran.bst: No hyphenation pattern has been}%
\typeout{** loaded for the language `#1'. Using the pattern for}%
\typeout{** the default language instead.}%
\else
\language=\csname l@#1\endcsname
\fi
#2}}

\bibitem{zhang2022ral}
L.~Zhang, M.~Helmberger, L.~F.~T. Fu, D.~Wisth, M.~Camurri, D.~Scaramuzza, and
  M.~Fallon, ``{HILTI-Oxford} dataset: A millimeter-accurate benchmark for
  simultaneous localization and mapping,'' \emph{IEEE Robot. Automat. Lett.},
  vol.~8, no.~1, pp. 408--415, 2022.

\bibitem{zhang2014rss}
J.~Zhang and S.~Singh, ``{LOAM: LiDAR odometry and mapping in real-time.}'' in
  \emph{Robot. Sci. Syst.}, vol.~2, no.~9, 2014, pp. 1--9.

\bibitem{lim2020normal}
H.~Lim, S.~Hwang, S.~Shin, and H.~Myung, ``Normal distributions transform is
  enough: Real-time {3D} scan matching for pose correction of mobile robot
  under large odometry uncertainties,'' in \emph{Proc. Int. Conf. Control,
  Automat. Syst.}, 2020, pp. 1155--1161.

\bibitem{lim2021ral}
H.~Lim, S.~Hwang, and H.~Myung, ``{ERASOR: Egocentric ratio of pseudo
  occupancy-based dynamic object removal for static 3D point cloud map
  building},'' \emph{IEEE Robot. Automat. Lett.}, vol.~6, no.~2, pp.
  2272--2279, 2021.

\bibitem{ren2019sensors}
Z.~Ren, L.~Wang, and L.~Bi, ``Robust {GICP-based 3D LiDAR SLAM} for underground
  mining environment,'' \emph{Sensors}, vol.~19, no.~13, pp. 2915--2933, 2019.

\bibitem{jung2020rs}
S.~Jung, D.~Choi, S.~Song, and H.~Myung, ``Bridge inspection using unmanned
  aerial vehicle based on {HG-SLAM: Hierarchical graph-based SLAM},''
  \emph{Remote Sens.}, vol.~12, no.~18, pp. 3022--3041, 2020.

\bibitem{sung2021isr}
C.~Sung, S.~Jeon, H.~Lim, and H.~Myung, ``{What if there was no revisit?
  Large-scale graph-based SLAM with traffic sign detection in an HD map using
  LiDAR inertial odometry},'' \emph{Intell. Serv. Robot.}, vol.~15, no.~2, pp.
  161--170, 2022.

\bibitem{song2022ral}
S.~Song, H.~Lim, A.~J. Lee, and H.~Myung, ``{DynaVINS: A visual-inertial SLAM
  for dynamic environments},'' \emph{IEEE Robot. Automat. Lett.}, vol.~7,
  no.~4, pp. 11\,523--11\,530, 2022.

\bibitem{kim2022ral-step}
Y.~Kim, B.~Yu, E.~M. Lee, J.-H. Kim, H.-W. Park, and H.~Myung, ``{STEP: State}
  estimator for legged robots using a preintegrated foot velocity factor,''
  \emph{IEEE Robot. Automat. Lett.}, vol.~7, no.~2, pp. 4456--4463, 2022.

\bibitem{noh2023ral}
D.~Noh, C.~Sung, T.~Uhm, W.~Lee, H.~Lim, J.~Choi, K.~Lee, D.~Hong, D.~Um,
  I.~Chung, \emph{et~al.}, ``{X-MAS}: Extremely large-scale multi-modal sensor
  dataset for outdoor surveillance in real environments,'' \emph{IEEE Robot.
  Automat. Lett.}, vol.~8, no.~2, pp. 1093--1100, 2023.

\bibitem{hu2022iros}
S.~Hu, Y.~Kim, H.~Lim, A.~J. Lee, and H.~Myung, ``{eCDT: Event} clustering for
  simultaneous feature detection and tracking,'' in \emph{Proc. IEEE/RSJ Int.
  Conf. Intell. Robot. Syst.}, 2022, pp. 3808--3815.

\bibitem{lee2022ral-vivid}
A.~J. Lee, Y.~Cho, Y.-S. Shin, A.~Kim, and H.~Myung, ``{ViViD++: Vision for
  visibility dataset},'' \emph{IEEE Robot. Automat. Lett.}, vol.~7, no.~3, pp.
  6282--6289, 2022.

\bibitem{lim2021ral-patch}
H.~Lim, M.~Oh, and H.~Myung, ``{Patchwork: Concentric zone-based region-wise
  ground segmentation with ground likelihood estimation using a 3D LiDAR
  sensor},'' \emph{IEEE Robot. Automat. Lett.}, vol.~6, no.~4, pp. 6458--6465,
  2021.

\bibitem{seo2022ur}
D.-U. Seo, H.~Lim, S.~Lee, and H.~Myung, ``{PaGO-LOAM: Robust ground-optimized
  LiDAR odometry},'' in \emph{Proc. Int. Conf. Ubiquti. Robot.}, 2022, pp.
  1--7.

\bibitem{burnett2021ral}
K.~Burnett, A.~P. Schoellig, and T.~D. Barfoot, ``{Do we need to compensate for
  motion distortion and doppler effects in spinning radar navigation?}''
  \emph{IEEE Robot. Automat. Lett.}, vol.~6, no.~2, pp. 771--778, 2021.

\bibitem{vizzo2023ral}
I.~Vizzo, T.~Guadagnino, B.~Mersch, L.~Wiesmann, J.~Behley, and C.~Stachniss,
  ``{KISS-ICP: In defense of point-to-point ICP -- Simple, accurate, and robust
  registration if done the right way},'' \emph{IEEE Robot. Automat. Lett.},
  2023, {DOI:} 10.1109/LRA.2023.3236571.

\bibitem{bai2022ral}
C.~Bai, T.~Xiao, Y.~Chen, H.~Wang, F.~Zhang, and X.~Gao, ``{Faster-LIO}:
  Lightweight tightly coupled {LiDAR-inertial} odometry using parallel sparse
  incremental voxels,'' \emph{IEEE Robot. Automat. Lett.}, vol.~7, no.~2, pp.
  4861--4868, 2022.

\bibitem{shan2018iros}
T.~Shan and B.~Englot, ``{LeGO-LOAM: Lightweight and ground-optimized LiDAR
  odometry and mapping on variable terrain},'' in \emph{Proc. IEEE/RSJ Int.
  Conf. Intell. Robot. Syst.}, 2018, pp. 4758--4765.

\bibitem{xu2021ral}
W.~Xu and F.~Zhang, ``{Fast-LIO: A fast, robust LiDAR-inertial odometry package
  by tightly-coupled iterated kalman filter},'' \emph{IEEE Robot. Automat.
  Lett.}, vol.~6, no.~2, pp. 3317--3324, 2021.

\bibitem{hinduja2019iros}
A.~Hinduja, B.-J. Ho, and M.~Kaess, ``Degeneracy-aware factors with
  applications to underwater {SLAM},'' in \emph{Proc. IEEE/RSJ Int. Conf.
  Intell. Robot. Syst.}, 2019, pp. 1293--1299.

\bibitem{li2022tim}
H.~Li, B.~Tian, H.~Shen, and J.~Lu, ``An intensity-augmented {LiDAR-inertial
  SLAM for solid-state LiDARs} in degenerated environments,'' \emph{IEEE Trans.
  Instrum. Meas.}, vol.~71, pp. 1--10, 2022.

\bibitem{du2023arxiv}
W.~Du and G.~Beltrame, ``Real-time simultaneous localization and mapping with
  {LiDAR} intensity,'' \emph{arXiv preprint arXiv:2301.09257}, 2023.

\bibitem{jiao2021icra}
J.~Jiao, Y.~Zhu, H.~Ye, H.~Huang, P.~Yun, L.~Jiang, L.~Wang, and M.~Liu,
  ``Greedy-based feature selection for efficient {LiDAR SLAM},'' in \emph{Proc.
  IEEE Int. Conf. Robot. Automat.}, 2021, pp. 5222--5228.

\bibitem{shan2021icra}
T.~Shan, B.~Englot, C.~Ratti, and D.~Rus, ``{LVI-SAM}: Tightly-coupled
  {LiDAR-visual-inertial} odometry via smoothing and mapping,'' in \emph{Proc.
  IEEE Int. Conf. Robot. Automat.}, 2021, pp. 5692--5698.

\bibitem{zhen2019icra}
W.~Zhen and S.~Scherer, ``Estimating the localizability in tunnel-like
  environments using {LiDAR} and {UWB},'' in \emph{Proc. IEEE Int. Conf. Robot.
  Automat.}, 2019, pp. 4903--4908.

\bibitem{zhou2021tvt}
H.~Zhou, Z.~Yao, and M.~Lu, ``{LiDAR/UWB} fusion based {SLAM} with
  anti-degeneration capability,'' \emph{IEEE Trans. Veh. Technol.}, vol.~70,
  no.~1, pp. 820--830, 2021.

\bibitem{park2017iccv}
C.~Park, S.~Kim, P.~Moghadam, C.~Fookes, and S.~Sridharan, ``Probabilistic
  surfel fusion for dense {LiDAR} mapping,'' in \emph{Proc. IEEE Int. Conf.
  Comput. Vis.}, 2017, pp. 2418--2426.

\bibitem{kang2016tro}
J.~Kang and N.~L. Doh, ``{Full-DOF calibration of a rotating 2-D LiDAR with a
  simple plane measurement},'' \emph{IEEE Trans. Robot.}, vol.~32, no.~5, pp.
  1245--1263, 2016.

\bibitem{wang2017tmi}
J.~Wang, R.~Schaffert, A.~Borsdorf, B.~Heigl, X.~Huang, J.~Hornegger, and
  A.~Maier, ``Dynamic {2-D/3-D} rigid registration framework using
  point-to-plane correspondence model,'' \emph{IEEE Trans. Med. Imaging.},
  vol.~36, no.~9, pp. 1939--1954, 2017.

\bibitem{schaffert2019tmi}
R.~Schaffert, J.~Wang, P.~Fischer, A.~Maier, and A.~Borsdorf, ``Robust
  multi-view {2-D/3-D} registration using point-to-plane correspondence
  model,'' \emph{IEEE Trans. Med. Imaging.}, vol.~39, no.~1, pp. 161--174,
  2019.

\end{thebibliography}
